\documentclass{article} 
\usepackage[preprint]{colm2026_conference}

\usepackage{microtype}
\usepackage{hyperref}
\usepackage{url}
\usepackage{amssymb} 
\usepackage{tabularx}
\usepackage{booktabs}
\usepackage{graphicx}
\usepackage{hyperref}
\usepackage{listings}
\usepackage{xcolor}
\usepackage{booktabs}
\usepackage{longtable}
\usepackage{subcaption}
\usepackage{array}
\usepackage{booktabs}
\usepackage{ragged2e}
\usepackage{subcaption}

\usepackage{lineno}

\definecolor{darkblue}{rgb}{0, 0, 0.5}
\hypersetup{colorlinks=true, citecolor=darkblue, linkcolor=darkblue, urlcolor=darkblue}

\title{ProjectionBench: Evaluating Scientific Hypothesis Generation in LLMs Under Progressive Information Disclosure}

\author{Andrew J.~Lew, Yuan Cao \& Markus J. Buehler \\
Unreasonable Labs \\
Mountain View, CA 94043, USA \\
\texttt{\{ajlew,yuancao,mbuehler\}@unreasonablelabs.ai} \\
}

\begin{document}

\ifcolmsubmission
\linenumbers
\fi

\maketitle

\begin{abstract}
Scientific discovery is an inherently creative and uncertain process, requiring reasoning beyond the recall of known knowledge. While many benchmarks have been proposed to evaluate large language model (LLM) performance on deep research tasks via multi-hop retrieval, their innovative reasoning abilities essential for true scientific discovery remain largely untested. We introduce a benchmark framework for evaluating model performance in scientific discovery and reasoning, building up from a raw problem to the classical null hypothesis test. In our framework, models initially receive only the topic and research question from a recent paper, with technical details progressively revealed. At each stage of information disclosure, the model is tasked with generating hypotheses that address the research question, which is compared with the conclusions from the original paper and evaluated via automated semantic similarity of constituent atomic claims. This progressive evaluation of semantic divergence from ground-truth conclusions enables assessment of a model’s innovativeness (under minimal information) to grounded reasoning capabilities (under full experimental details), both critical for using LLMs for scientific discovery purposes. Our framework provides a foundation for systematically evaluating scientific reasoning and discovery capabilities in LLMs, crucial for advancing the development of next-generation AI scientist/co-scientist systems. Specifically, here we evaluate GPT-5, GPT-5.4, Gemini 2.5 pro, and Gemini 3.1 pro preview across 45 papers spanning bioactive materials, mechanical materials, and nanomaterials. We find that GPT-5.4 and Gemini 3.1 pro outperform their previous generation counterparts as expected, and GPT-5.4 in particular maintains 0.7 F1 score alignment with ground truth conclusions even under minimal context. 
\end{abstract}

\section{Introduction}
Using large language models (LLMs) for scientific discovery and knowledge creation has emerged as an important frontier \cite{VanNoorden2023AIScience, Eger2025TransSci, Reddy2025SciDisc}. However, how to rigorously evaluate model performance on such tasks remains an open challenge. Addressing this requires first clarifying what constitutes discovery. At its core, discovery involves identifying something previously unknown, and go beyond the retrieval of existing knowledge or interpolation within known patterns. Yet novelty alone is insufficient: a valid discovery must also be grounded in and logically consistent with established knowledge. In an era of growing concern over the quality and reliability of AI-generated content \cite{Shaib2025AISlop}, it is therefore critical to distinguish meaningful discoveries from unsupported or erroneous claims. Creativity without correctness does not constitute discovery, it constitutes misinformation.

Despite a growing body of work on LLM evaluation, systematic benchmarks for measuring scientific discovery remain limited. Existing benchmarks have primarily focused on factual recall, knowledge retrieval, and reasoning within well-defined or exam-style contexts. For instance, SciBench \cite{Wang2024SciBench} evaluates LLMs on college-level problems in physics, chemistry, and mathematics, revealing only moderate success rates even for state-of-the-art systems. Similarly, MatSciBench 
\cite{Zhang2025MatSciBench} conducts evaluation on over 1000 college-level textbook questions across subdisciplines of materials science. More broadly, ResearcherBench 
\cite{Xu2025ResearcherBench} examines deep research information retrieval capabilities on expert curated questions through coverage, faithfulness, and groundedness scores evaluating the completeness and appropriateness of citations. DeepScholar-Bench \cite{Patel2025DeepScholarBench} similarly grades retrieved information, and does so via knowledge synthesis, retrieval quality, and verifiability scores of generated related work summaries. And ScholarEval \cite{Moussa2025ScholarEval} assesses a model’s ability to contextualize research ideas against retrieved literature, measuring both soundness (whether analogous methods have succeeded in the literature) and contribution (defined as the extent of advancement along dimensions such as data, methods, evaluation, and conceptual framework used). Other efforts such as DiscoveryBench \cite{Majumder2024DiscoveryBench} seek to evaluate how good current state-of-the-art LLMs are at automated data-driven discovery, specifically by testing model ability to generate high specificity hypotheses given a dataset and discovery goal. And InnovatorBench \cite{Wu2025InnovatorBench} moves closer to measuring end-to-end scientific innovation by benchmarking models on 20 tasks derived from influential LLM research papers, though its reliance on curated tasks and multi-day execution makes it resource-intensive.

However, the very presence of a dataset already implies some prior assumptions about potential outcomes, reflected in what data was collected and how it was collected. During earlier stages of scientific exploration, researchers may not yet have a dataset if they are still in the experimental ideation phase. At this stage, a researcher might have only a general topic and research question, and may be considering how a proposed experiment could shape or refine their further understanding of natural relationships. Despite progress in evaluation schemes, existing benchmarks primarily emphasize literature retrieval, citation alignment, or specific intensive analysis tasks; a lightweight, general probe of scientific discovery ability remains largely missing.

To address this gap, we propose a benchmarking framework that evaluates an LLM’s capacity for scientific discovery by building up to the classical structure of null hypothesis testing. Because both hypothesis innovativeness and reasoning accuracy are central to scientific discovery, the benchmark is designed to capture how model behavior evolves across different levels of contextual information. Specifically, it spans settings from a minimal configuration - where only the topic and research question is provided - to a maximal configuration that includes the ground truth null hypothesis and experimental procedures. Given papers never observed by the model before, a strong model for scientific discovery should be able to propose novel hypotheses that approximate ground-truth conclusions from a raw research question, and derive accurate conclusions when additional experimental context is introduced.

While many existing benchmarks evaluate LLMs on tasks involving scientific reasoning, they typically rely on established textbook knowledge, expert annotations, or comparisons against retrieved literature. In contrast, benchmarks designed to assess discovery capabilities often depend on heavily curated scenarios, limiting their scalability and preventing continuous, real-time updates. In this work, we aim to evaluate both reasoning and discovery capabilities in a live, scalable setting. Table \ref{tab:table1} provides a high-level comparison of representative benchmarks across key dimensions, including data sources, prediction targets, evaluation methodologies, and their support for reasoning, discovery, live evaluation, and scalability.

\begin{table}[t]
\centering
\small
\setlength{\tabcolsep}{3pt}
\caption{Comparison across related scientific benchmarks (R=Reasoning, D=Discovery, L=Live, S=Scalable)}\label{tab:table1}
\begin{tabularx}{\textwidth}{l X X X c c c c}
\toprule
\textbf{Benchmark} & \textbf{Data Source} & \textbf{Target} & \textbf{Evaluation} & \textbf{R} & \textbf{D} & \textbf{L} & \textbf{S} \\
\midrule
SciBench (\citeyear{Wang2024SciBench}) & College textbooks & QA & Ref sol. & $\checkmark$ & X & X & $\checkmark$ \\

DiscoveryBench (\citeyear{Majumder2024DiscoveryBench}) & Expert tasks & Task impl. & LLM judge & $\checkmark$ & $\checkmark$ & X & X \\

DiscoveryWorld (\citeyear{Jansen2024DiscoveryWorld}) & Expert tasks & Task impl. & Ref + LLM/human & $\checkmark$ & $\checkmark$ & X & X \\

MatSciBench (\citeyear{Zhang2025MatSciBench}) & College textbooks & QA & Ref + LLM & $\checkmark$ & X & X & $\checkmark$ \\

ResearcherBench (\citeyear{Xu2025ResearcherBench}) & Expert Qs & QA & Human + LLM & $\checkmark$ & $\checkmark$ & X & X \\

FrontierScience (\citeyear{Wand2025FrontierScience}) & Expert Qs & QA & LLM judge & $\checkmark$ & X & X & X \\

DeepScholar-Bench (\citeyear{Patel2025DeepScholarBench}) & Papers & Related work & LLM judge & $\checkmark$ & X & $\checkmark$ & $\checkmark$ \\

ScholarEval (\citeyear{Moussa2025ScholarEval}) & Papers & Ideation & LLM judge & $\checkmark$ & X & $\checkmark$ & $\checkmark$ \\

InnovatorBench (\citeyear{Wu2025InnovatorBench}) & Expert tasks & Task impl. & Ref sol. & $\checkmark$ & $\checkmark$ & X & X \\

AIRS-Bench (\citeyear{lupidi2026airsbenchsuitetasksfrontier}) & Papers & Task impl & Ref sol. &$\checkmark$ & $\checkmark$ & X & X \\

ResearchGym (\citeyear{garikaparthi2026researchgymevaluatinglanguagemodel}) & Papers &  Task impl. & Task metrics & $\checkmark$ & $\checkmark$ & X & X \\

InnoEval (\citeyear{qiao2026innoevalresearchideaevaluation}) & Papers & Ideation & LLM judge & X & \checkmark & X & X \\

\toprule

\textbf{ProjectionBench (Ours)} & Papers & Outcomes & LLM judge & $\checkmark$ & $\checkmark$ & $\checkmark$ & $\checkmark$ \\
\bottomrule
\end{tabularx}
\end{table}

The remainder of this paper is organized as follows. Section 2 defines the discovery task and describes the null hypothesis-based prompting procedure. Section 3 introduces a granular automated grading methodology. Section 4 details the live dataset curation process focused on recent materials science publications. Section 5 presents comparative results across several state-of-the-art models, and Section 6 concludes with a discussion of implications, limitations, and directions for future work in benchmarking model-driven scientific discovery.

\section{Discovery Task}
Here, we consider the task of scientific discovery within the classic framework of the null hypothesis test. Namely, that experiments “exist only in order to give the facts a chance of disproving the null hypothesis” (Fisher, 1935). In this emulation of building up to null hypothesis testing, we expose models-under-test to a two-element prompt. In the first element, we present a variable amount of research context to the model. In the second element, we ask for the model to provide its best projection on the solution or findings of the research question. By varying the amount of research context from just the raw topic and research question, to including a base null hypothesis, and then including a specific experimental procedure, we test the full gamut of model ability from de novo scientific discovery to structured reasoning. The specific framework used is in the Appendix as Prompt 1.     

An ideal model for scientific discovery should combine strong creativity with rigorous reasoning. A highly creative model should be able to generate novel insights even with minimal context, while a strong reasoning model should accurately infer conclusions when experimental procedures are provided. Our benchmark is designed to capture both ends of this spectrum, enabling a comprehensive assessment of a model’s capability in scientific research.

\section{Automated Grading Method}
\subsection{Granular Grading Approach}
Once a model has projected the results of an experimental procedure, we can compare that projection to known ground truth in order to grade its performance. A canonical formulation of scientific hypotheses concentrates on “if/then” or “when/then” style relationships between independent and dependent variables \cite{Elsevier2024Hypothesis}. However, when comparing the content of projected and ground truth results, statements could be composed of multiple such relationships depending on the complexity of the research question (e.g. If $X$, then $Y$ and $Z$). 

Thus, here we first break down the potentially complex results of a scientific study into its atomic claims, with each claim formatted as a relationship between independent and dependent variables. Specifically, independent variables are the conditions of an experiment that are systematically manipulated by the investigator and dependent variables are the outcomes that are measured in an experiment as a result of experimental manipulation of the independent variables. We break down a scientific result statement into its constituent claims by leveraging GPT-5 as an evaluatory model, using Prompt 2 in the Appendix.

Once we have the ground truth result broken up into its constituent relationship claims, we can analogously extract the relationship claims from the projected result. We utilize a similar prompt for this extraction, only now providing the previous list of ground truth claims and prompting for the extraction of analogous statements, using Prompt 3 in the Appendix.

Finally, as there is no guarantee that the claims of the projected result map one-to-one with the claim of the ground truth result, we use Prompt 4 in the Appendix to extract remaining extraneous claims from the projected result that do not have a direct analogue to the ground truth. 

\begin{figure}[t]
    \centering
    \includegraphics[width=\linewidth]{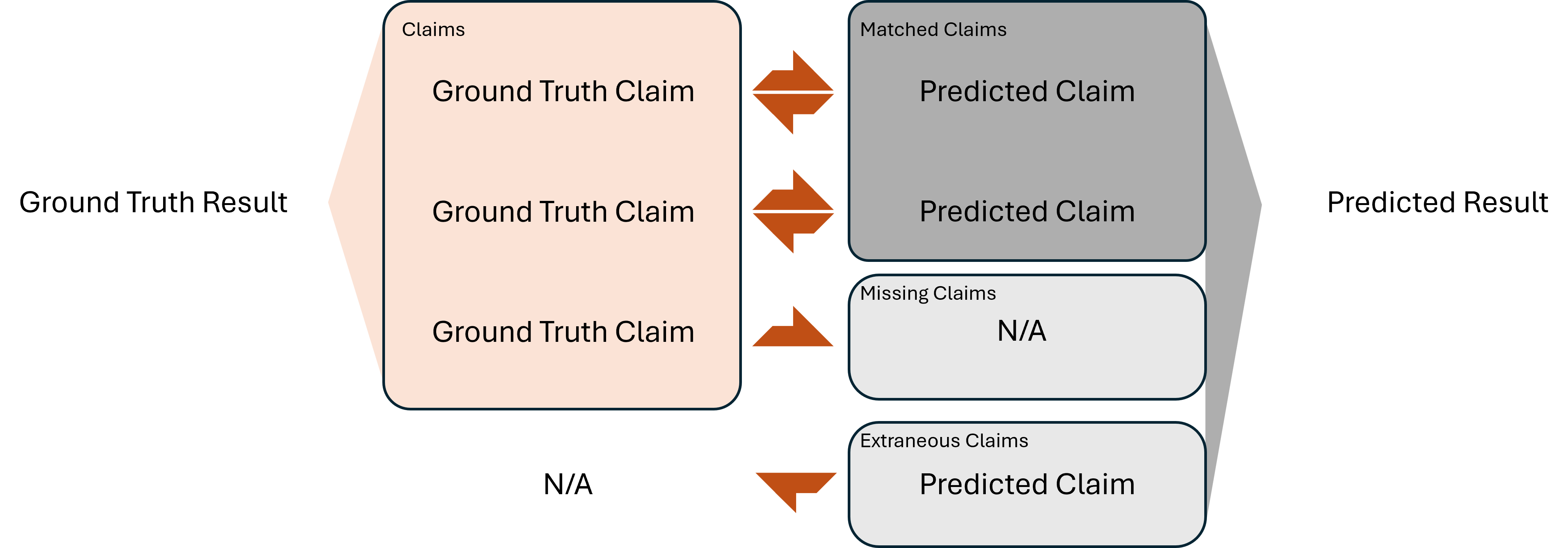}
    \caption{Ground truth and projected results can be broken up into their constituent claims for more granular comparison. Generally, imperfect projections may miss aspects of the ground truth, or include extraneous claims beyond the ground truth.}
    \label{fig:figure1}
\end{figure}

To summarize, our comparison of projected and ground truth results is visualized in Figure \ref{fig:figure1} Each result statement is broken down into their constituent claims, with three potential cases to consider: a) a ground truth claim is paired with a projected claim, b) a ground truth claim is missing a corresponding projected claim, and c) an extraneous projected claim does not have a matching ground truth claim. 

\subsection{Calculating Grade Score}
As the components of the projection may not map one-to-one onto the ground truth, we can assess how well the projection performs by measuring its precision and recall, and then combine these two measures into a single grade using the F1 score. To do so, we first evaluate if each projected claim is a true positive, false positive, or inconclusive - again leveraging GPT-5 as an evaluatory model. Specifically, we use GPT-5 in an LLM-as-a-judge \cite{Gu2025LLMJudge} capacity to compare the alignment between ground truth and projected claims with the Prompt 5 rubric in the Appendix. Note that all projections from ground truth and models-under-test are passed through GPT-5 for claim extraction, such that all alignment comparisons are done on outputs filtered through GPT-5. This is to lessen biases that may arise from the judge model potentially favoring its own phrasing styles over those of other models. In order to address potential concerns over positional bias \cite{Shi2025JudgingJudges} we run the alignment scoring twice, flipping the order of the ground truth and projected claims, and average the alignment scores if the evaluations differ.

For each projected claim with a paired ground truth claim, we can simply calculate the alignment score between paired claims. However, for each extraneous projected claim without a corresponding ground truth, we instead compare to each ground truth claim and take the average score to measure its overall impact. True positives are defined as the magnitude of positive scores—projected claims that align with ground truth—as shown in Equation~\ref{eq:tp}:

\begin{equation}
\mathrm{TP}(g,p) = 
\sum_{i=1}^{M} \max\big(a(g_i, p_i), 0\big)
+ 
\sum_{j=M+1}^{N} \max\left( \frac{1}{M} \sum_{i=1}^{M} a(g_i, p_j), 0 \right)
\label{eq:tp}
\end{equation}

False positives are defined as the magnitude of negative scores—projected claims that contradict ground truth—as shown in Equation~\ref{eq:fp}:

\begin{equation}
\mathrm{FP}(g,p) = 
- \sum_{i=1}^{M} \min\big(a(g_i, p_i), 0\big)
- 
\sum_{j=M+1}^{N} \min\left( \frac{1}{M} \sum_{i=1}^{M} a(g_i, p_j), 0 \right)
\label{eq:fp}
\end{equation}

The total number of relevant elements is defined as all positive elements—the sum of all ground truth claims and all positive extraneous claims—as shown in Equation~\ref{eq:re}:

\begin{equation}
\mathrm{RE}(g,p) = 
M + 
\sum_{j=M+1}^{N} \mathbf{1} \left\{
\frac{1}{M} \sum_{i=1}^{M} a(g_i, p_j) > 0
\right\}
\label{eq:re}
\end{equation}

where $g$ denotes the ground truth, $p$ denotes the projection, $M$ is the number of ground truth claims, $N$ is the total number of claim categories (ground truth plus extraneous projections), and $a(g,p)$ is the LLM-as-a-judge alignment score between ground truth and projection.

Precision, recall, and F1 score are computed in the standard manner. To comprehensively assess a model's scientific discovery and reasoning capabilities, we adopt the Area Under the Curve (AUC) as the final evaluation metric, defined as the aggregate of F1 scores across varying disclosure levels.

\subsection{Validating Grading Approach}
\begin{figure}[t]
    \centering
    \includegraphics[width=0.8\linewidth]{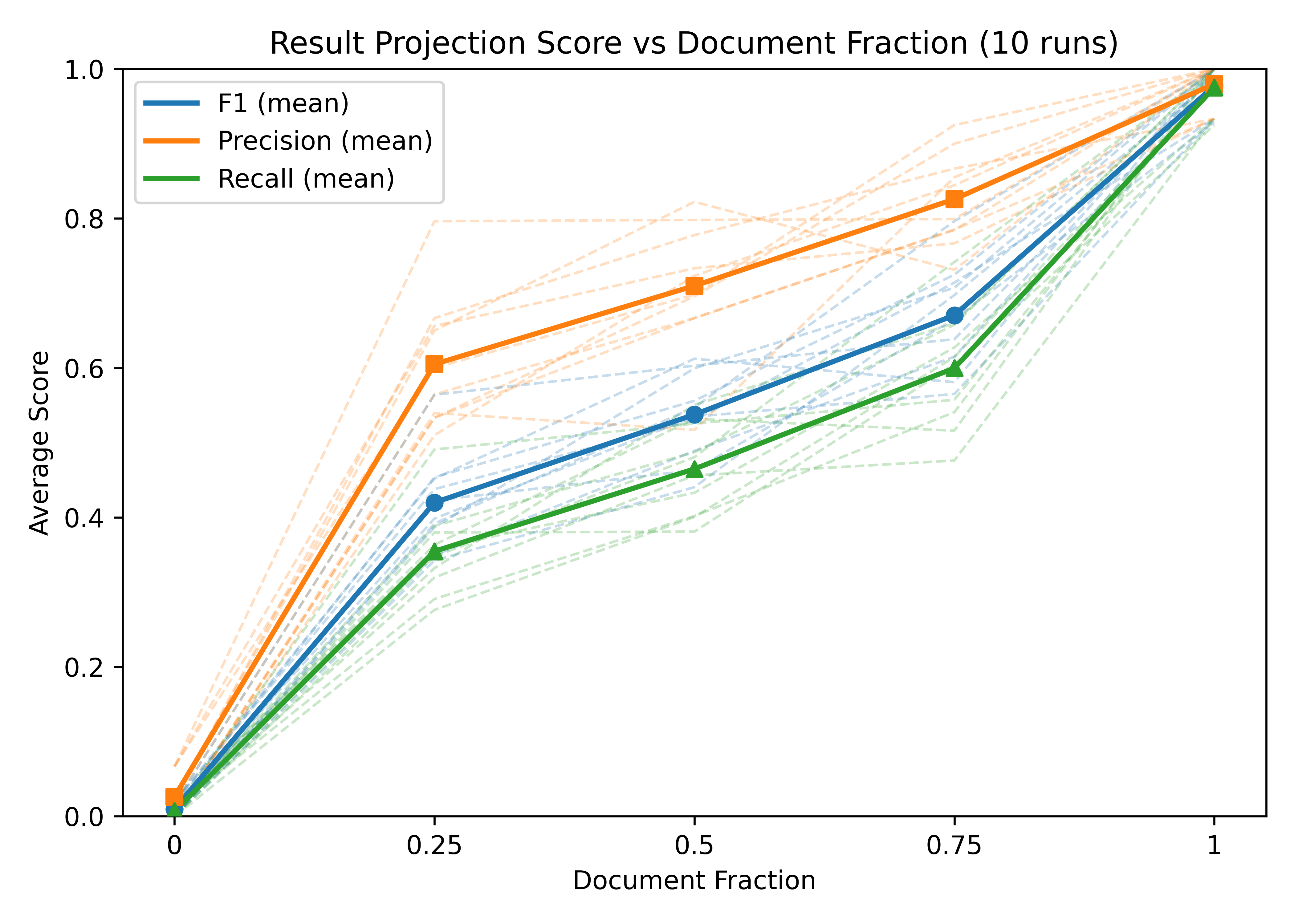}
    \caption{Scoring procedure is aligned with fraction of ground truth document provided, using GPT-5.}
    \label{fig:figure2}
\end{figure}

Importantly, this grading method must be well calibrated to real claim alignment in order to be valid. To verify the correctness of the scoring scheme, we conduct a series of tests using toy claims that have controlled overlap with the true ground truth. These validation tests use GPT-5 throughout, so as to control potential differences in models and only measure the behavior of the scoring procedure itself. Specifically, we extract these toy claims from randomly sampled internal slices of each ground truth manuscript, with varying lengths. Toy claims obtained from larger fractions of the manuscript should approach the ground truth - and yield scores approaching 1 (up to some limit of LLM stochasticity) for the full manuscript. Vice versa, toy claims extracted from less provided manuscript should approach 0 (down to the limit of no manuscript and no claims). Fractional amounts of provided manuscript should yield claims (and thus scores) correlated to the fraction of provided information. Figure \ref{fig:figure2} confirms the expected behavior over an $N$ of 10 runs. 

\section{Dataset Curation}
To benchmark the ability for different models-under-test to conduct scientific discovery in this rapidly evolving field, we design a live procedure continuously updating from Open Access Springer Nature articles published within the last 6 months. This is to ensure manuscripts are published after the training cutoff dates for models-under-test \cite{HaoooWang2025Cutoff} in order to avoid training contamination. Specifically, here we show results using scientific studies obtained with the Open Access Springer Nature API with search terms “bioactive materials”, “nanomaterials”, and “mechanical materials”, with 15 manuscripts per category for a total of 45 manuscripts, detailed in the Appendix Table \ref{tab:manuscripts}. The benchmark will be released at \href{https://github.com/<future-link>}{https://github.com/\textlangle future-link\textrangle }.

To prepare these manuscripts for testing, we parse their full-text with GPT-5 to extract the following key pieces of information in three waves: 1) Title, Topic, Experimental Procedure, 2) Hypothesis, 3) Null Hypothesis, Research Question, and Results.

Note the following order in which we extract this information:
We start with broad factual pieces with more limited room for interpretation, such as title, topic, and experimental procedure.

We next extract the hypothesis given the already extracted title, topic, and experimental procedure as context. We use this extraction order to better constrain the extracted hypotheses to fit the main substance of the study. Otherwise, attempting to extract a hypothesis without context may yield varying results - e.g. the extracted hypothesis may incorrectly incorporate elements of ancillary introduction or proposed future applications, instead of focusing on the main topic tested by the experimental. 

With the core hypotheses and experimental procedure extracted, we use these as context for extracting the corresponding null hypothesis, the initial research question, and the final results. This procedure encourages that all extracted elements are logically connected to each other, and that the extracted results address the research question via rejecting the null hypothesis by experiment. 

The exact prompts for information extraction are shown in the Appendix Prompt 6. 

\section{Experiment and Results}

We evaluate several state-of-the-art models on the task of predicting the outcomes of scientific studies, including GPT-5, GPT-5.4, Gemini 2.5 pro, and Gemini 3.1 pro preview. Specifically, we structure the information extracted from the dataset curation outlined in Section 4 into the payloads for the discovery task described in Section 2. As a key measure for preventing contamination of projected results by simply retrieving the correct answers to each research question, we conduct all model responses in offline mode. After yielding projected results for each manuscript, for each model, for each variable context amount (just the raw topic and research question, adding the null hypothesis, and adding the experimental procedure), we score against the extracted ground truth results according to the grading method outlined in Section 3, with scores shown in Figure \ref{fig:figure3} and detailed in the Appendix Table \ref{tab:scores}. The overall AUC of these models are given in Table \ref{tab:model_auc}.

\begin{table}[h]
\caption{Performance of models in projecting scientific outcomes measured by AUC.}\label{tab:model_auc}
\centering
\begin{tabular}{l c}
\hline
\textbf{Model} & \textbf{AUC} \\
\hline
GPT-5 & 1.44 \\
GPT-5.4 & 1.56 \\
Gemini 2.5 Pro & 1.33 \\
Gemini 3.1 Pro Preview & 1.44 \\
\hline
\end{tabular}

\end{table}

\begin{figure}[t]
    \centering
    \begin{subfigure}[t]{0.48\textwidth}
        \centering
        \includegraphics[width=\linewidth]{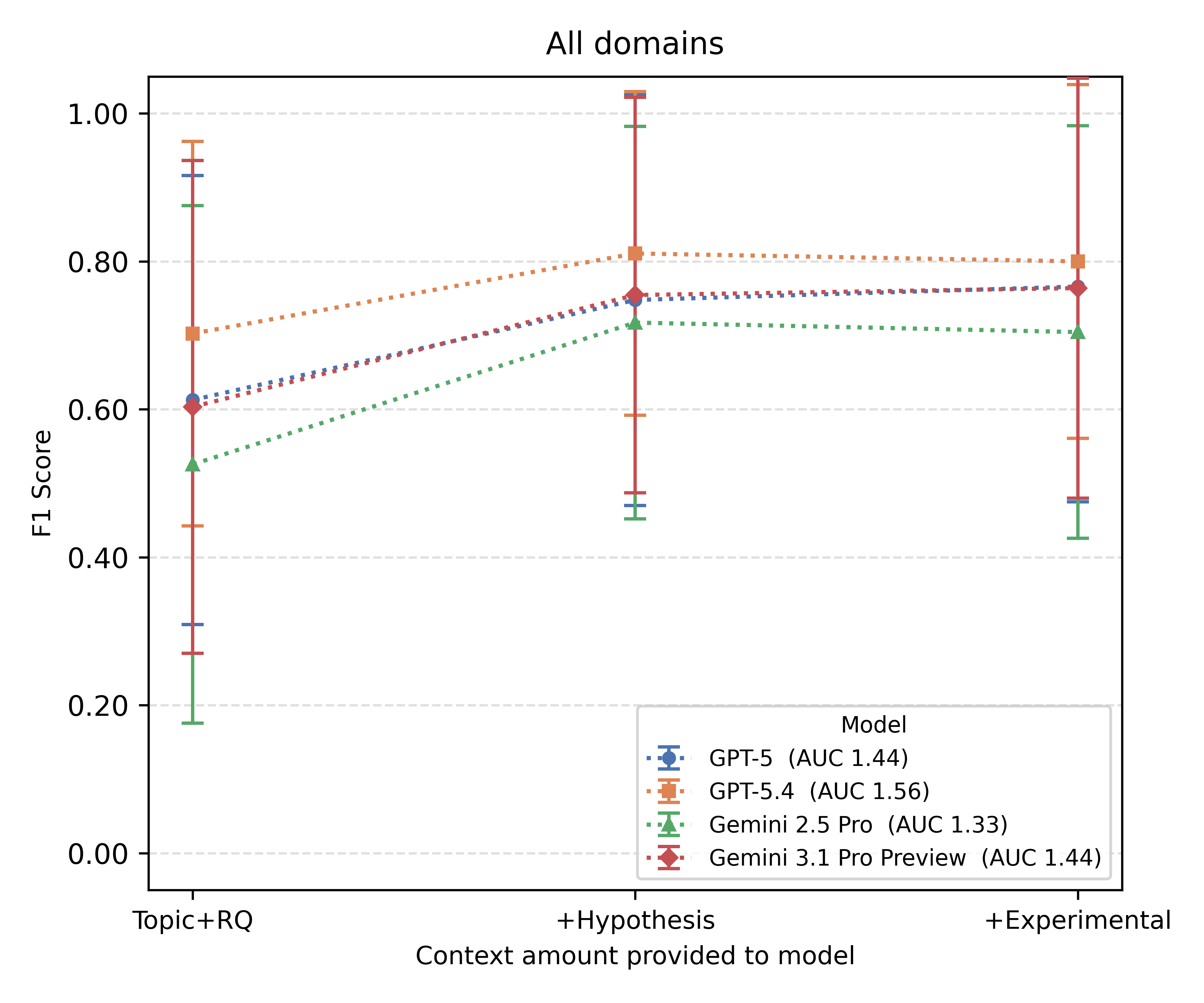}
        \caption{}
        \label{fig:figure3a}
    \end{subfigure}
    \hfill
    \begin{subfigure}[t]{0.48\textwidth}
        \centering
        \includegraphics[width=\linewidth]{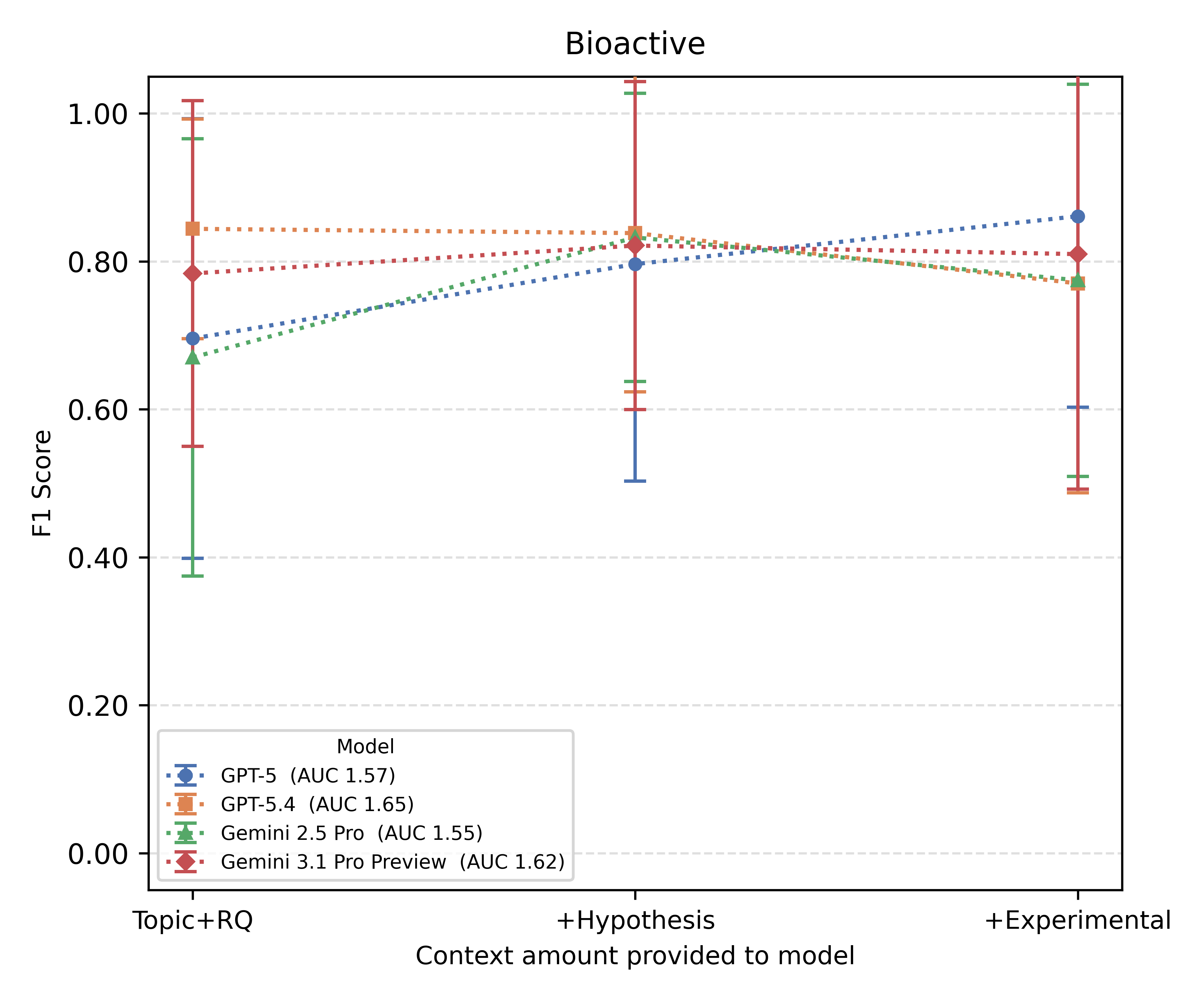}
        \caption{}
        \label{fig:figure3b}
    \end{subfigure}

    \vspace{0.5em}

    \begin{subfigure}[t]{0.48\textwidth}
        \centering
        \includegraphics[width=\linewidth]{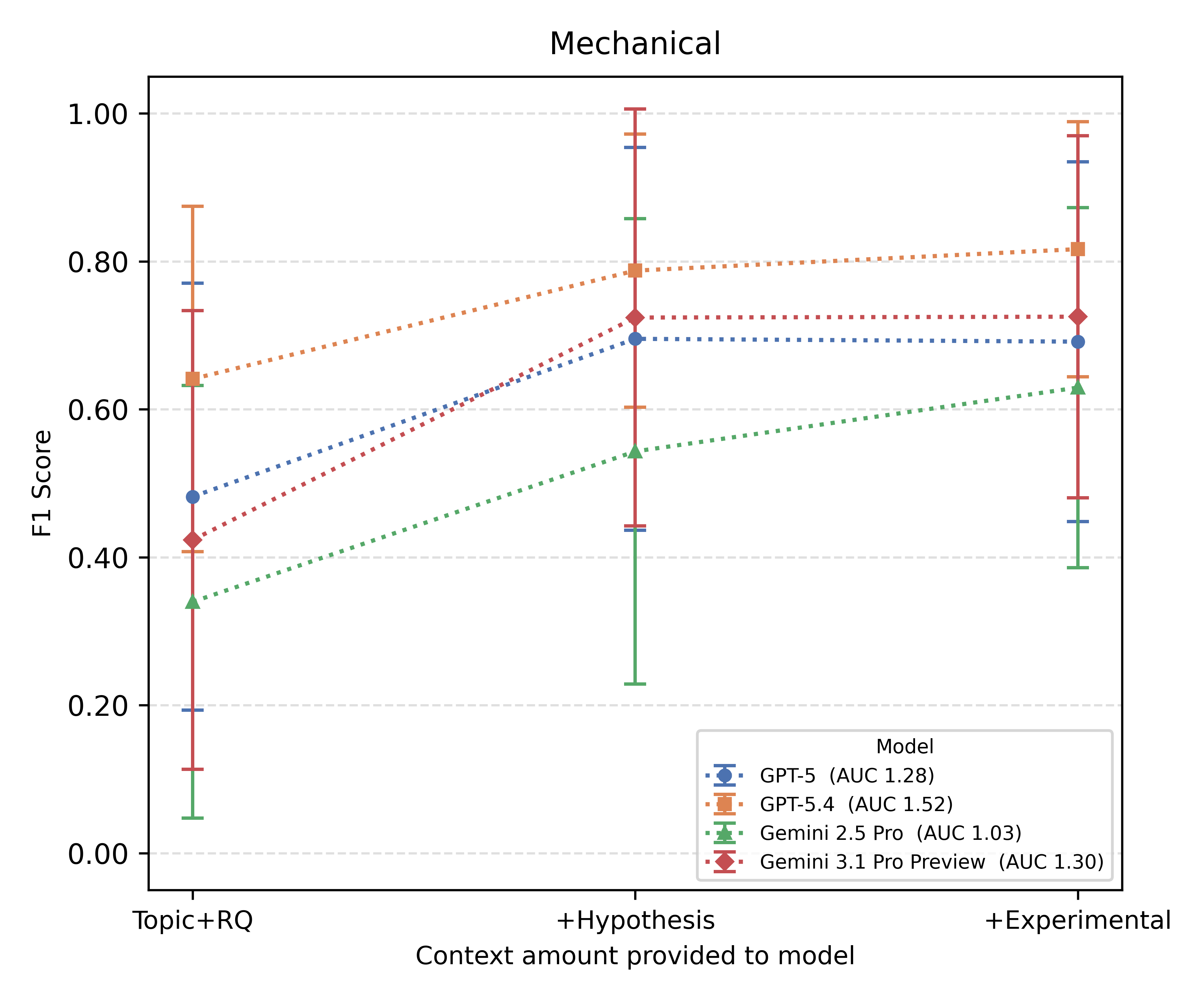}
        \caption{}
        \label{fig:figure3c}
    \end{subfigure}
    \hfill
    \begin{subfigure}[t]{0.48\textwidth}
        \centering
        \includegraphics[width=\linewidth]{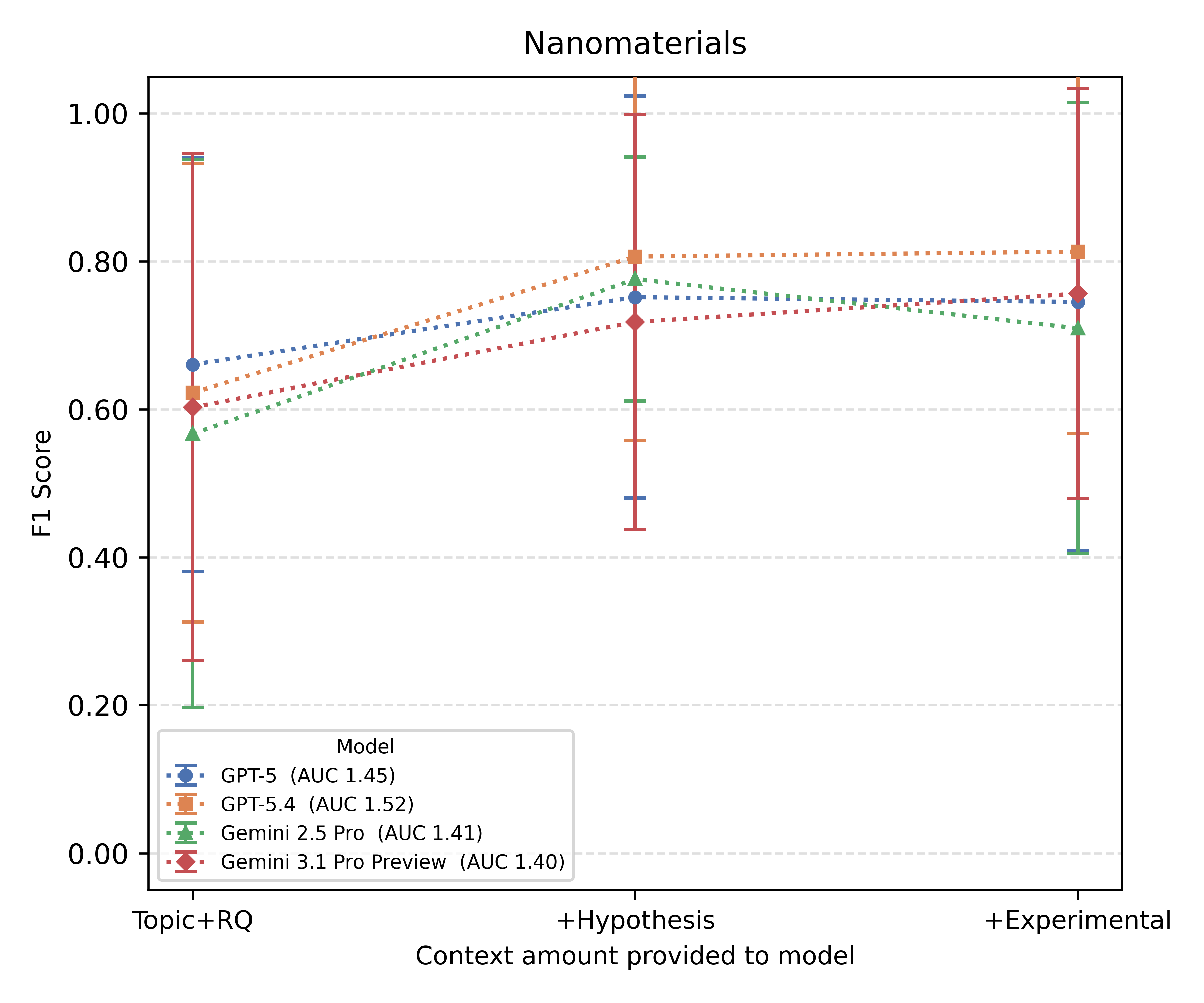}
        \caption{}
        \label{fig:figure3d}
    \end{subfigure}

    \caption{(a) AUC score comparisons across models-under-test and amount of context provided. (b)--(d) Scores separated by domain category (bioactive, mechanical, nanomaterials).}
    \label{fig:figure3}
\end{figure}

As shown in the results, the more recent GPT-5.4 and Gemini 3.1 Pro Preview models consistently outperform their earlier counterparts across all levels of provided context. Performance generally improves with additional context. However augmenting the raw topic and research question with a null hypothesis yields a larger gain than further specifying an experimental procedure on top of that hypothesis, indicating diminishing marginal returns from increasingly detailed guidance.

Notably, GPT-5.4 maintains a relatively strong performance (F1 $\approx$ 0.70) even under minimal context, whereas the earlier Gemini 2.5 Pro model requires more complete experimental details to achieve comparable results. Across domains, bioactive discovery tasks yield higher scores than mechanical and nanomaterial categories, with smaller relative improvements from added context. This pattern suggests either an imbalance in the underlying knowledge distribution of frontier models or early saturation in reasoning performance for these domains.

Finally, in mechanical manuscripts, Gemini 3.1 Pro Preview surpasses GPT-5 under high-context conditions that emphasize structured reasoning, but exhibits slightly lower performance in low-context settings, which rely more heavily on open-ended discovery. This contrast highlights a potential trade-off between structured reasoning and exploratory generation capabilities across models.

\begin{figure}[t]
    \centering
    \includegraphics[width=\linewidth]{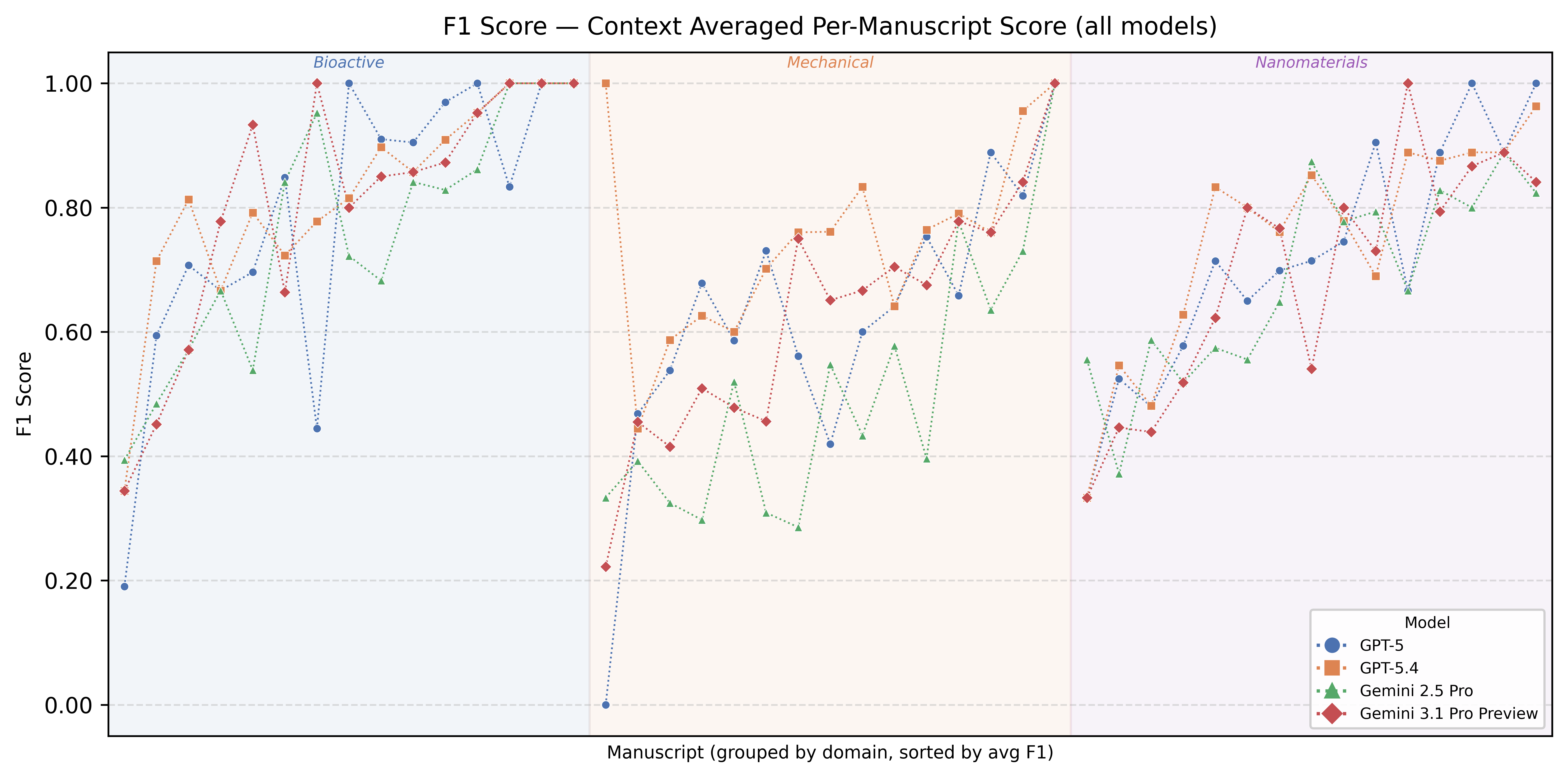}
    \caption{Context averaged per-manuscript scores illustrate broad range of difficulties in projecting scientific discoveries}
    \label{fig:figure4}
\end{figure}

It is important to note that the standard deviations across all scores are relatively large. This variability arises from the heterogeneous difficulty of the manuscripts included in the evaluation set. As illustrated in Figure \ref{fig:figure4}, certain manuscripts are substantially more challenging to predict than others. In particular, bioactive manuscripts tend to cluster near the upper performance bound, suggesting a saturation effect. In contrast, mechanical manuscripts exhibit a much wider distribution of scores, indicating greater variability and overall lower predictability. Nanomaterial manuscripts do not display the same saturation behavior as bioactive ones, but they maintain a higher performance floor compared to mechanical manuscripts. These observations suggest that current frontier models capture certain aspects of bioactive knowledge relatively well, while mechanical domains remain less well-modeled and offer significant room for improvement. Nanomaterials appear to occupy an intermediate position, where baseline knowledge is present but accurate, high-confidence predictions remain challenging.

An illustrative example of model projections is shown in Table \ref{tab:comparison}. Here, we show a brief comparison of results for the mechanical manuscript "Enhancing thermo-mechanical properties of Honckenya fiber-reinforced polypropylene composites: a comparative study of novel potash salt and NaOH treatments" (10.1038/s41598-026-43630-5). We can see that GPT-5.4 provides a partially aligned answer even under minimal context conditions - getting the main relationships correct but spuriously projecting an optimum level when one was not described in the ground truth. Under the ability to reason over the full experimental details, this changes to projecting all the same relationships as the ground truth, just with less confident verbiage. In contrast, at low context Gemini 2.5 pro incorrectly projects that the conventional NaOH treatment is better, indicating its anchor to historical knowledge rather than an innovative projection. When presented with more context, Gemini 2.5 pro is able to capture the storage modulus and thermal stability effects, but misses the loss modulus effect and still maintains that the effects are only comparable to conventional NaOH treatment, rather than projecting the ground truth improvement. We can see how ProjectionBench evaluates not only a model's ability to reason out a scientific outcome, but also how well it can innovate and reach valid new findings.

\begin{table}[ht]
\small
\centering
\caption{Example projection comparison}\label{tab:comparison}
\begin{tabular}{
    >{\RaggedRight\arraybackslash}p{1.5cm} 
    >{\RaggedRight\arraybackslash}p{5.5cm} 
    >{\RaggedRight\arraybackslash}p{5.5cm} 
}
\toprule
\textbf{} & \textbf{Topic + RQ} & \textbf{with Hypothesis + Experimental} \\
\midrule

\textbf{Ground Truth} &
\multicolumn{2}{c}{\parbox[t]{11cm}{This study finds KTN-treated Honckenya fiber/polypropylene composites exhibit significantly \textbf{higher storage and loss moduli} and \textbf{improved thermal stability} over temperature than NaOH-treated or untreated counterparts.\newline}} 
\\[12pt]

\textbf{GPT-5.4} & 
This study finds potash treatment produces comparable to modestly \textbf{superior thermo-mechanical performance than NaOH} in natural fiber polymer composites, with tensile/flexural properties, \textbf{storage modulus, and thermal stability improving} with treatment \textbf{up to an optimum level and then declining} under harsher treatment. \newline\textit{F1 Score: 0.5}\newline &
This study finds KTN-treated Honckenya fiber/polypropylene composites exhibit \textbf{higher storage and loss moduli} and modestly \textbf{improved thermal stability} across the temperature range than untreated composites, with thermo-mechanical performance that is comparable to or slightly \textbf{better than NaOH}-treated composites. \newline\textit{F1 Score: 1.0}\newline \\[12pt]

\textbf{Gemini 2.5 pro} & 
This study finds that while potash treatment imparts superior thermal stability to the composite, the \textbf{conventional NaOH treatment yields a more significant improvement} in mechanical strength. \newline\textit{F1 Score: 0.0}\newline &
This study finds that natural potash salt (KTN) treatment \textbf{enhances the storage modulus and thermal stability} of Honckenya fiber composites to a level \textbf{comparable with conventional NaOH} treatment, with both significantly outperforming composites made with untreated fibers.
\newline\textit{F1 Score: 0.8}\newline \\

\bottomrule
\end{tabular}
\end{table}

\section{Conclusions}
Scientific discovery is inherently creative, uncertain, and forward-looking. Unlike recall-based benchmarks that evaluate known knowledge through exam-style questions, anticipating the outcomes of novel research - and assessing a model’s ability to generate genuinely new insights - poses a fundamentally more challenging problem, requiring nuanced reasoning and deep domain understanding.

In this work, we propose an initial framework for benchmarking a model’s capacity to project scientific outcomes. Drawing inspiration from null hypothesis testing, our approach evaluates how effectively large language models can reason toward plausible scientific results rather than simply retrieve them. In high-context settings, the framework measures a model’s ability to synthesize and reason over rich, structured inputs; in low-context settings, it assesses the model's capacity to generate valid and innovative projections from minimally specified research questions. These results also provide practical guidance on how the amount of context provided may influence model performance.

Future work will focus on expanding the benchmark across broader scientific domains, incorporating a wider range of models, and refining evaluation protocols—for example, by varying context granularity and incorporating deeper analysis of reasoning processes. We also acknowledge a potential limitation in our current evaluation setup: using GPT-5 as the judge for GPT-family models may introduce bias. To address this, future iterations will incorporate cross-family evaluation using independent models. Ultimately, we position this benchmark as an early step toward more rigorous evaluation of machine-assisted discovery, and as a guidepost for the development of systems that can more meaningfully accelerate scientific research.

\bibliography{colm2026_conference}

@misc{Eger2025TransSci,
  title={Transforming Science with Large Language Models: A Survey on AI-assisted Scientific Discovery, Experimentation, Content Generation, and Evaluation},
  author={Steffen Eger and Yong Cao and João D'Souza and Atticus Geiger and Christian Greisinger and Sabine Gross and Yifan Hou and Brigitte Krenn and Anne Lauscher and Ying Li and Chenghua Lin and Nafise Sadat Moosavi and Wei Zhao and Tim Miller},
  year={2025},
  eprint={2502.05151},
  archivePrefix={arXiv},
  primaryClass={cs.AI},
  url={https://arxiv.org/abs/2502.05151}
}

@misc{Gu2025LLMJudge,
  title={A Survey on LLM-as-a-Judge},
  author={Jiaqi Gu and Xiaoyan Jiang and Zhiyuan Shi and Haotian Tan and Xiaohui Zhai and Chao Xu and Wei Li and Yujia Shen and Shanshan Ma and Hao Liu and Shuai Wang and Kai Zhang and Zichao Lin and Bo Zhang and Li Ni and Wei Gao and Yu Wang and Jia Guo},
  year={2025},
  eprint={2411.15594},
  archivePrefix={arXiv},
  primaryClass={cs.CL},
  url={https://arxiv.org/abs/2411.15594}
}

@misc{HaoooWang2025Cutoff,
  author = {Hao Wang},
  title = {LLM Knowledge Cutoff Dates},
  year = {2025},
  howpublished = {\url{https://github.com/HaoooWang/llm-knowledge-cutoff-dates}},
  note = {GitHub repository}
}

@misc{Jansen2024DiscoveryWorld,
  title={DiscoveryWorld: A Virtual Environment for Developing and Evaluating Automated Scientific Discovery Agents},
  author={Peter Jansen and Michael A. Cote and Tushar Khot and Eric Bransom and Bhavana D. Mishra and Bodhisattwa Prasad Majumder and Oyvind Tafjord and Peter Clark},
  year={2024},
  eprint={2406.06769},
  archivePrefix={arXiv},
  primaryClass={cs.AI},
  url={https://arxiv.org/abs/2406.06769}
}

@misc{Majumder2024DiscoveryBench,
  title={DiscoveryBench: Towards Data-Driven Discovery with Large Language Models},
  author={Bodhisattwa Prasad Majumder and Harsh Surana and Deepak Agarwal and Bhavana D. Mishra and Ankit Meena and Aman Prakhar and Tanmay Vora and Tushar Khot and Ashish Sabharwal and Peter Clark},
  year={2024},
  eprint={2407.01725},
  archivePrefix={arXiv},
  primaryClass={cs.AI},
  url={https://arxiv.org/abs/2407.01725}
}

@misc{Moussa2025ScholarEval,
  title={SCHOLAREVAL: Research Idea Evaluation Grounded in Literature},
  author={Hady N. Moussa and Pedro Q. Da Silva and Daniel Adu-Ampratwum and Andrew East and Zhiyong Lu and Niccolò Puccetti and Ming Xue and Huan Sun and Bodhisattwa Prasad Majumder and Saurabh Kumar},
  year={2025},
  eprint={2510.16234},
  archivePrefix={arXiv},
  primaryClass={cs.CL},
  url={https://arxiv.org/abs/2510.16234}
}

@misc{Patel2025DeepScholarBench,
  title={DeepScholar-Bench: A Live Benchmark and Automated Evaluation for Generative Research Synthesis},
  author={Lalit Patel and Negar Arabzadeh and Hritik Gupta and Aravind Sundar and Ion Stoica and Matei Zaharia and Carlos Guestrin},
  year={2025},
  eprint={2508.20033},
  archivePrefix={arXiv},
  primaryClass={cs.IR},
  url={https://arxiv.org/abs/2508.20033}
}

@misc{Reddy2025SciDisc,
  title={Towards Scientific Discovery with Generative AI: Progress, Opportunities, and Challenges},
  author={Chandan K. Reddy and Parshin Shojaee},
  year={2025},
  eprint={2412.11427},
  archivePrefix={arXiv},
  primaryClass={cs.AI},
  url={https://arxiv.org/abs/2412.11427}
}

@misc{Shaib2025AISlop,
  title={Measuring AI "Slop" in Text},
  author={Chadi Shaib and Tuhin Chakrabarty and David Garcia-Olano and Byron C. Wallace},
  year={2025},
  eprint={2509.19163},
  archivePrefix={arXiv},
  primaryClass={cs.CL},
  url={https://arxiv.org/abs/2509.19163}
}

@misc{Shi2025JudgingJudges,
  title={Judging the Judges: A Systematic Study of Position Bias in LLM-as-a-Judge},
  author={Linyuan Shi and Chao Ma and Weixin Liang and Xinyu Diao and Wei Ma and Soroush Vosoughi},
  year={2025},
  eprint={2406.07791},
  archivePrefix={arXiv},
  primaryClass={cs.CL},
  url={https://arxiv.org/abs/2406.07791}
}

@misc{Elsevier2024Hypothesis,
  author = {Elsevier},
  title = {Step-by-Step Guide: Crafting a Strong Hypothesis},
  year = {2024},
  howpublished = {\url{https://scientific-publishing.webshop.elsevier.com/manuscript-preparation/what-how-write-good-hypothesis-research/}},
  note = {Elsevier Author Services}
}

@article{VanNoorden2023AIScience,
  author = {Richard Van Noorden and Jeffrey M. Perkel},
  title = {AI and science: what 1,600 researchers think},
  journal = {Nature},
  volume = {621},
  number = {7980},
  pages = {672--675},
  year = {2023}
}

@misc{Wang2024SciBench,
  title={SciBench: Evaluating College-Level Scientific Problem-Solving Abilities of Large Language Models},
  author={Xiaofei Wang and Zhiruo Hu and Pan Lu and Yizhong Zhu and Jiacheng Zhang and Siva Subramaniam and Aditya R. Loomba and Sheng Zhang and Yiming Sun and Wei Wang},
  year={2024},
  eprint={2307.10635},
  archivePrefix={arXiv},
  primaryClass={cs.AI},
  url={https://arxiv.org/abs/2307.10635}
}

@misc{Wand2025FrontierScience,
  title={FrontierScience: Evaluating AI's Ability to Perform Expert-level Scientific Tasks},
  author={Michael Wand and Rui Lin and Kai Hu and Jian Jiao and Niloofar Chowdhury and Eric Chang and Tanishq Patwardhan},
  year={2025},
  eprint={2601.21165},
  archivePrefix={arXiv},
  primaryClass={cs.AI},
  url={https://arxiv.org/abs/2601.21165}
}

@misc{Wu2025InnovatorBench,
  title={InnovatorBench: Evaluating Agents' Ability to Conduct Innovative LLM Research},
  author={Yifan Wu and Dan Fu and Wenhao Si and Zhen Huang and Ming Jiang and Kai Li and Shuo Xia and Jing Sun and Tianyi Xu and Xia Hu and Pan Lu and Xiaoyang Cai and Long Le and Wei Zhu and Yang Xiao and Peng Liu},
  year={2025},
  eprint={2510.27598},
  archivePrefix={arXiv},
  primaryClass={cs.AI},
  url={https://arxiv.org/abs/2510.27598}
}

@misc{Xu2025ResearcherBench,
  title={ResearcherBench: Evaluating Deep AI Research Systems on the Frontiers of Scientific Inquiry},
  author={Tianyi Xu and Pan Lu and Lifan Ye and Xia Hu and Peng Liu},
  year={2025},
  eprint={2507.16280},
  archivePrefix={arXiv},
  primaryClass={cs.AI},
  url={https://arxiv.org/abs/2507.16280}
}

@misc{Zhang2025MatSciBench,
  title={MatSciBench: Benchmarking the Reasoning Ability of Large Language Models in Materials Science},
  author={Jiawei Zhang and Junjie Gan and Xiaofei Wang and Zhen Jia and Chen Gu and Jie Chen and Yizhong Zhu and Ming D. Ma and Dong Zhou and Lei Li and Wei Wang},
  year={2025},
  eprint={2510.12171},
  archivePrefix={arXiv},
  primaryClass={cond-mat.mtrl-sci},
  url={https://arxiv.org/abs/2510.12171}
}

@misc{lupidi2026airsbenchsuitetasksfrontier,
      title={AIRS-Bench: a Suite of Tasks for Frontier AI Research Science Agents}, 
      author={Alisia Lupidi and Bhavul Gauri and Thomas Simon Foster and Bassel Al Omari and Despoina Magka and Alberto Pepe and Alexis Audran-Reiss and Muna Aghamelu and Nicolas Baldwin and Lucia Cipolina-Kun and Jean-Christophe Gagnon-Audet and Chee Hau Leow and Sandra Lefdal and Hossam Mossalam and Abhinav Moudgil and Saba Nazir and Emanuel Tewolde and Isabel Urrego and Jordi Armengol Estape and Amar Budhiraja and Gaurav Chaurasia and Abhishek Charnalia and Derek Dunfield and Karen Hambardzumyan and Daniel Izcovich and Martin Josifoski and Ishita Mediratta and Kelvin Niu and Parth Pathak and Michael Shvartsman and Edan Toledo and Anton Protopopov and Roberta Raileanu and Alexander Miller and Tatiana Shavrina and Jakob Foerster and Yoram Bachrach},
      year={2026},
      eprint={2602.06855},
      archivePrefix={arXiv},
      primaryClass={cs.AI},
      url={https://arxiv.org/abs/2602.06855}, 
}

@misc{garikaparthi2026researchgymevaluatinglanguagemodel,
      title={ResearchGym: Evaluating Language Model Agents on Real-World AI Research}, 
      author={Aniketh Garikaparthi and Manasi Patwardhan and Arman Cohan},
      year={2026},
      eprint={2602.15112},
      archivePrefix={arXiv},
      primaryClass={cs.AI},
      url={https://arxiv.org/abs/2602.15112}, 
}

@misc{qiao2026innoevalresearchideaevaluation,
      title={InnoEval: On Research Idea Evaluation as a Knowledge-Grounded, Multi-Perspective Reasoning Problem}, 
      author={Shuofei Qiao and Yunxiang Wei and Xuehai Wang and Bin Wu and Boyang Xue and Ningyu Zhang and Hossein A. Rahmani and Yanshan Wang and Qiang Zhang and Keyan Ding and Jeff Z. Pan and Huajun Chen and Emine Yilmaz},
      year={2026},
      eprint={2602.14367},
      archivePrefix={arXiv},
      primaryClass={cs.CL},
      url={https://arxiv.org/abs/2602.14367}, 
}
\bibliographystyle{colm2026_conference}

\appendix
\section{Appendix}
\subsection*{Prompt 1: Scientific Projection Task}
\begin{lstlisting}
def context_amount(input_data, amount):
    base = f'Topic: {input_data['topic']} \nResearch Question: {input_data['research_question']}'
    if amount == 0:
        return base
    with_hypothesis = base + f'\nUnverified Hypothesis: {input_data['null_hypothesis_ground_truth']}'
    if amount == 1:
        return with_hypothesis
    with_method = with_hypothesis + f'\nExperimental Procedure: {input_data['experimental_method']}'
    if amount == 2:
        return with_method

element1_prompt = context_amount(input_data, amount)

element2_prompt = f'''
In one sentence, do your best to project the key outcome of the Research Question.
Focus on the existence and qualitative extent of relationships between Independent and Dependent Variables.
Do not explain the problem or method. Only provide the new result observation as a statement.
Provide in the following format, filling in 'RESULT':

This study finds RESULT
'''

total_prompt = element1_prompt + element2_prompt
\end{lstlisting}

\subsection*{Prompt 2: Ground Truth Claim Extraction}

\begin{lstlisting}
json_template = '''{'claims': [
    {
    'subject': '[Independent Variable]',
    'relationship': '[no more than 5 words]',
    'object': '[Dependent Variable]'
    }
]
}'''

analyze_claims_prompt = f'''
Independent Variables: The conditions of an experiment that are systematically manipulated by the investigator.
Dependent Variables: The outcomes that are measured in an experiment as a result of an experimental manipulation of the independent variable(s).

Relationship Claims: What are the qualitative relationships between Independent and Dependent Variables?

Break down the following Passage into the key Relationship Claims it makes. Be MECE.
Format your answer as JSON. If a value cannot be filled, leave as '':
{json_template}

Passage:
{passage}
'''
\end{lstlisting}

\subsection*{Prompt 3: Analogous Claim Extraction}

\begin{lstlisting}
json_template = json.dumps(previous_claims)

analyze_analogous_claims_prompt = f'''
Independent Variables: The conditions of an experiment that are systematically manipulated by the investigator.
Dependent Variables: The outcomes that are measured in an experiment as a result of an experimental manipulation of the independent variable(s).

Relationship Claims: What are the qualitative relationships between Independent and Dependent Variables?

Here is a set of Example claims.
Example claims:
{json_template}

Use the following Passage to extract new Relationship Claims analogous to the Example claims. Be mutually exclusive.
Format your answer as JSON. If a value cannot be filled, report as ''.
Each 'relationship' should only be limited to 5 words.
You must do your best to give an analogous claim matching topics for each of the Example claims.
If an analogous claim is unavailable, then report as 'N/A' in the JSON values. 

Passage:
{passage}
'''
\end{lstlisting}

\subsection*{Prompt 4: Remaining Claim Extraction}

\begin{lstlisting}
json_template = json.dumps(previous_claims)

prompt = f'''
Independent Variables: The conditions of an experiment that are systematically manipulated by the investigator.
Dependent Variables: The outcomes that are measured in an experiment as a result of an experimental manipulation of the independent variable(s).

Relationship Claims: What are the qualitative relationships between Independent and Dependent Variables?

We have started a Draft breaking down the target Passage into the Relationship Claims it makes. 
Draft claims:
{json_template}

Does this Draft miss key Relationship Claims made by the following target Passage? Be mutually exclusive.
If so, report up to 3 unique missing Relationship Claims in JSON format. Else, simply return an empty JSON as {json.dumps({'claims': []})}.
Format your answer as JSON. If a value cannot be filled, leave as '':

Passage:
{passage}
'''
\end{lstlisting}

\subsection*{Prompt 5: Alignment Rubric}

\begin{lstlisting}
alignment_prompt = f'''
You are an expert in logically evaluating the alignment between two scientific statements.
You will be given two items to compare and must evaluate their alignment based on the following rubric:

-1. Misaligned: The two items indicate opposing relationships, reporting differing directions for an effect such as '[A] increases' vs '[A] decreases', '[A] decreases' vs '[A] is unchanged', '[A] is unaffected' vs '[A] increases', etc.
0. Uncorrelated: The two items neither support nor contradict each other, reporting different effects such as '[A] increases' vs '[B] increases', '[A] increases' vs '[B] decreases', '[A] is unchanged' vs '[B] decreases', etc. 
1. Aligned: The two items indicate supporting relationships, reporting similar directions for an effect such as '[A] greatly increases' vs '[A] markedly increases', '[A] is unchanged' vs '[A] is not affected', '[A] is suppressed' vs '[A] is limited', etc.

In your response, first provide a brief rationale for your rating, then clearly state the numeric rating (-1 to 1) according to the rubric above in the format 'RATING: [number]'.

Item 1: {prediction}
Item 2: {ground_truth}
'''
\end{lstlisting}

\subsection*{Prompt 6: Document Parsing}

\begin{lstlisting}
role_message = 'You are an expert analyst specializing in clear, objective science communication.'

# --- Wave 1 ---
title_extraction = f'''{role_message}
Scientific Passage: {paragraph}

What is the title of the study in the given Scientific Passage?
'''

topic_extraction = f'''{role_message}
Scientific Passage: {paragraph}

In less than 5 words, what is the main general topic of this Scientific Passage?
Provide just the topic as a short phrase.
'''

experimental_extraction = f'''{role_message}
Scientific Passage: {paragraph}

In one paragraph, describe the experimental steps conducted in this Scientific Passage.
Only report the methods as a numbered list of objective physical instructions. 
Do not include any results. Do not express subjective qualifications.
Provide in the tone of a factual, terse, statement of procedures, starting with:

1.) 
'''

# --- Wave 2 ---
hypothesis_extraction = f'''{role_message}
Scientific Passage: {paragraph}
Experimental Procedure: {experimental}

In one brief objective sentence, what was the main verifiable/falsifiable hypothesis tested by the Experimental Procedure? 
Focus on qualitative relationships between variables.
Do not explain the problem or method. Only provide the Hypothesis as a statement.
Provide in the following format - choose 'If' or 'When' where appropriate, then fill in 'CONDITION' and 'OBSERVABLE':

(If/When) CONDITION then OBSERVABLE.
'''

# --- Wave 3 ---
null_hypothesis_extraction = f'''{role_message}
Topic: {topic}
Alternative Hypothesis: {hypothesis} 

In one brief objective sentence, formulate the corresponding Null Hypothesis to the given Alternative Hypothesis.
Focus on qualitative relationships between variables. Be unbiased in language, such as 'no change' instead of 'no increase' or 'no decrease'. 
Do not explain the problem or method. Only provide the Null Hypothesis as a statement.
Provide in the following format - choose 'If' or 'When' where appropriate, then fill in 'CONDITION' and 'OBSERVABLE':

(If/When) CONDITION then OBSERVABLE.
'''

question_extraction = f'''{role_message}
Scientific Passage: {paragraph}
Hypothesis: {hypothesis} 

In less than 10 words, what is the scientific research question probed by the Hypothesis? 
Provide just the broad open-ended question, no leading details or answers.
'''

result_extraction  = f'''{role_message}
Scientific Passage: {paragraph}
Hypothesis: {hypothesis} 
Experimental Procedure: {experimental}

In one sentence, what was the new key insight observed as a result of this Scientific Passage? 
Focus on the existence and qualitative extent of each relationship in the Hypothesis.
Do not explain the problem or method. Only provide the new observation as a statement.
Provide in the following format, filling in 'RESULT':

This study finds RESULT
'''
\end{lstlisting}

\begin{longtable}{>{\raggedright\arraybackslash}p{2.5cm} >{\raggedright\arraybackslash}p{5.5cm} >{\centering\arraybackslash}p{1.3cm} >{\centering\arraybackslash}p{2.7cm}}
\caption{Manuscripts from Springer Nature Open Access API}\label{tab:manuscripts} \\
\toprule
DOI & Title & Date & Category \\
\midrule
\endfirsthead

\toprule
DOI & Title & Date & Category \\
\midrule
\endhead

10.1186/s12935-026-04276-5 & Ferroptosis targeting: a novel therapeutic armamentarium in multiple myeloma & 3/25/2026 & bioactive materials \\
10.1186/s44365-026-00033-x & Bioactive peptides in aquaculture: a peptidomics approach to food security & 3/12/2026 & bioactive materials \\
10.1038/s41598-026-44943-1 & ZIF-8 functionalized PCL/BG composite scaffolds with improved bioactivity and osteogenic differentiation & 3/20/2026 & bioactive materials \\
10.1007/s00339-026-09407-3 & Fabrication of silicate-based bioactive glass-containing sodium alginate hydrogel microfibers for tissue engineering applications & 3/26/2026 & bioactive materials \\
10.1186/s12870-026-08517-7 & Integrated high-density genetic maps in multi-parent F2 mapping populations provide a framework to unravel the genomic basis of agro-qualitative and metabolic traits in pepper (Capsicum annuum) & 3/9/2026 & bioactive materials \\
10.1007/s44463-026-00061-0 & Goat milk-derived bioactive peptides as a health promoter: an insight based on recent in silico studies & 3/10/2026 & bioactive materials \\
10.1186/s13023-026-04266-w & Assessing the nutritional value and health risks of special low-protein foods: narrative review & 3/6/2026 & bioactive materials \\
10.1007/s00216-026-06412-6 & Development of a metabolomic LC-QTOF method for bioactive compound production control in plant cell cultures & 3/14/2026 & bioactive materials \\
10.1007/s00339-026-09330-7 & Multifunctional coatings based on copper-doped bioactive glass/gelatin using sol-gel technique for stainless steel 316 L implant & 3/13/2026 & bioactive materials \\
10.1186/s12951-026-04172-0 & Coupled mesoporous silica nanoparticles and limonene-chitosan Pickering emulsions: enhanced insecticidal delivery and selectivity & 3/19/2026 & bioactive materials \\
10.1007/s44340-025-00047-6 & Marine macroalgal metabolites in microbial modulation for next-generation microecological therapeutics & 3/9/2026 & bioactive materials \\
10.1007/s44187-026-00925-w & Current trends in food safety analysis and quality control for food tablets & 3/11/2026 & bioactive materials \\
10.1038/s41598-026-43765-5 & Nutritional and regional assessment of wild anardana (Punica granatum L.) genotypes from the Pir Panjal range with implications for genetic resource utilization & 3/18/2026 & bioactive materials \\
10.1007/s11694-026-04142-y & Nutraceutical biochromes in grain crop matrices & 3/5/2026 & bioactive materials \\
10.1007/s44447-026-00136-w & Enhancing the recovery of bioactive compounds from rain tree (Samanea saman) seeds through optimized ultrasound-assisted extraction and response surface methodology & 3/12/2026 & bioactive materials \\

10.1186/s12890-026-04252-9 & Tracheostomy timing and weaning outcomes following prolonged mechanical ventilation & 3/26/2026 & mechanical materials \\
10.1038/s41598-026-43630-5 & Enhancing thermo-mechanical properties of Honckenya fiber-reinforced polypropylene composites: a comparative study of novel potash salt and NaOH treatments & 3/24/2026 & mechanical materials \\
10.1038/s41598-026-44474-9 & Impact of Alumina-Graphene nanoplatelets on the microstructure, corrosion behaviour, mechanical properties of cast aluminium nanocomposite by Vortex technique & 3/26/2026 & mechanical materials \\
10.1007/s00339-026-09427-z & Optimizing phase formation of CMR perovskite La$_{0.67}$Sr$_{0.33}$MnO$_3$ through grinding time & 3/26/2026 & mechanical materials \\
10.1007/s42114-026-01752-4 & Composition-driven phase structures in laser-deposited titanium-steel composites: microstructural evolution and interfacial strengthening mechanisms & 3/24/2026 & mechanical materials \\
10.1007/s44174-026-00692-3 & Integrated Mechanical Testing and Finite Element Modelling of a Biomorphic Bioceramic Scaffold for Spinal Fusion Applications & 3/24/2026 & mechanical materials \\
10.1007/s40948-025-01028-z & Mechanical properties of Indonesian organic-rich shales from the Central Sumatra and North Sumatra Basin, and comparison with U.S. shales & 3/26/2026 & mechanical materials \\
10.1186/s42252-026-00092-2 & An emerging liquid thermoplastic resin based structural composite through VARTM: a comprehensive comparison of the mechanical performance with conventional epoxy based composites at ambient and elevated temperature & 3/26/2026 & mechanical materials \\
10.1186/s40069-026-00913-5 & Effect of Recycled Ceramic Fine Aggregate (RCFA) on Strength and Durability of Concrete with Varying Water-To-Binder Ratios & 3/24/2026 & mechanical materials \\
10.1007/s00170-026-17882-4 & Experimental and analytical investigation of microstructure homogenization, internal defect repair, and material softening in LPBF-printed AlSi10Mg aluminum alloy via surface friction stir treatment & 3/24/2026 & mechanical materials \\

10.1186/s12989-026-00672-x & The effects of ingested cellulose nanomaterials on DNA methylation in intestinal cells & 3/12/2026 & nanomaterials \\
10.1007/s43939-026-00634-2 & Metal and carbon-based nanoparticles for environmental remediation: performance metrics, reaction mechanisms, and ecotoxicological implications & 3/29/2026 & nanomaterials \\
10.1186/s12951-026-04281-w & Nanomaterials for Alzheimer’s disease: emerging strategies in diagnosis and therapy & 3/21/2026 & nanomaterials \\
10.1186/s40712-026-00438-5 & The effect of base concentration on titanium dioxide nanotubes properties synthesized by hydrothermal method & 3/14/2026 & nanomaterials \\
10.1007/s00210-026-05170-7 & Rudolf Buchheim Award 2026: Toward a ToxAtlas of carbon-based nanomaterials: single-cell RNA sequencing reveals initiating cell circuits in pulmonary inflammation & 3/18/2026 & nanomaterials \\
10.1007/s00216-026-06443-z & Validated workflows for preparing and characterizing core-stained and surface-labeled fluorescent polymer particles with simple commercial automation tools & 3/18/2026 & nanomaterials \\
10.1007/s10971-026-07116-0 & Green synthesis of Mn$_3$O$_4$ nanomaterials by Cassia tora leaves extract for supercapacitor device with activated carbon & 3/24/2026 & nanomaterials \\
10.1186/s13065-026-01757-6 & Development of a novel voltammetric method for revefenacin determination using modified electrodes in pharmaceutical and biological matrices & 3/27/2026 & nanomaterials \\
10.1007/s44397-026-00050-4 & Recent advances in luminescent chemosensors for sensitive and selective detection of heavy metal ions in aqueous environments & 3/12/2026 & nanomaterials \\
10.1186/s40712-026-00419-8 & A review of novel metal oxide nanomaterials sensing mechanisms and synthesis techniques for gas sensing applications & 3/12/2026 & nanomaterials \\
10.1007/s10971-025-07023-w & Sol Gel synthesis of highly efficient Mn doped ZnO nanostructures for the effective photocatalytic degradation of organic pollutants & 3/13/2026 & nanomaterials \\
10.1007/s12551-026-01428-9 & Nano-based spectroscopic approaches for early diagnosis of Alzheimer’s disease: critical insights into amyloid-$\beta$ and tau biomarker biology and detection tools & 3/19/2026 & nanomaterials \\
10.1186/s44316-026-00053-6 & Versatile applications of EPS secreting microbes: advances in microbial fuel cells and beyond & 3/26/2026 & nanomaterials \\
10.1007/s44291-026-00188-w & Design and characterization of ZnO/NiO and ZnO/CuO heterostructures for high-performance solar cell applications & 3/15/2026 & nanomaterials \\
10.1186/s12870-026-08441-w & Optimizing biomass production and antioxidant dynamics in Ceratophyllum demersum L. using multi-walled carbon nanotubes and machine learning models & 3/11/2026 & nanomaterials \\

\bottomrule
\end{longtable}

\begin{longtable}{>{\raggedright\arraybackslash}p{2.5cm} >{\raggedright\arraybackslash}p{2.5cm} >{\raggedright\arraybackslash}p{2.5cm} >{\centering\arraybackslash}p{1cm}>{\centering\arraybackslash}p{1cm}>{\centering\arraybackslash}p{1.5cm}}
\caption{Scoring results}\label{tab:scores} \\
\toprule
Model & Category & Context Amount & F1 Mean & F1 Std & AUC \\
\midrule
\endfirsthead

\toprule
Model & Category & Context Amount & F1 Mean & F1 Std & AUC \\
\midrule
\endhead

GPT-5 & All & Topic+RQ & 0.6127 & 0.3035 & 1.43715 \\
GPT-5 & All & Hypothesis & 0.7478 & 0.2781 & \\
GPT-5 & All & Experimental & 0.7660 & 0.2910 & \\

GPT-5 & Bioactive & Topic+RQ & 0.6957 & 0.2973 & 1.5747 \\
GPT-5 & Bioactive & Hypothesis & 0.7963 & 0.2933 & \\
GPT-5 & Bioactive & Experimental & 0.8611 & 0.2580 & \\

GPT-5 & Mechanical & Topic+RQ & 0.4819 & 0.2888 & 1.2820 \\
GPT-5 & Mechanical & Hypothesis & 0.6953 & 0.2587 & \\
GPT-5 & Mechanical & Experimental & 0.6915 & 0.2433 & \\

GPT-5 & Nanomaterials & Topic+RQ & 0.6604 & 0.2800 & 1.45455 \\
GPT-5 & Nanomaterials & Hypothesis & 0.7517 & 0.2718 & \\
GPT-5 & Nanomaterials & Experimental & 0.7453 & 0.3365 & \\

GPT-5.4 & All & Topic+RQ & 0.7024 & 0.2599 & 1.56185 \\
GPT-5.4 & All & Hypothesis & 0.8107 & 0.2186 & \\
GPT-5.4 & All & Experimental & 0.7999 & 0.2391 & \\

GPT-5.4 & Bioactive & Topic+RQ & 0.8440 & 0.1484 & 1.6453 \\
GPT-5.4 & Bioactive & Hypothesis & 0.8382 & 0.2147 & \\
GPT-5.4 & Bioactive & Experimental & 0.7702 & 0.2829 & \\

GPT-5.4 & Mechanical & Topic+RQ & 0.6411 & 0.2335 & 1.5163 \\
GPT-5.4 & Mechanical & Hypothesis & 0.7875 & 0.1846 & \\
GPT-5.4 & Mechanical & Experimental & 0.8165 & 0.1724 & \\

GPT-5.4 & Nanomaterials & Topic+RQ & 0.6223 & 0.3096 & 1.5240 \\
GPT-5.4 & Nanomaterials & Hypothesis & 0.8063 & 0.2488 & \\
GPT-5.4 & Nanomaterials & Experimental & 0.8131 & 0.2458 & \\

Gemini 2.5 Pro & All & Topic+RQ & 0.5256 & 0.3498 & 1.3322 \\
Gemini 2.5 Pro & All & Hypothesis & 0.7172 & 0.2653 & \\
Gemini 2.5 Pro & All & Experimental & 0.7044 & 0.2788 & \\

Gemini 2.5 Pro & Bioactive & Topic+RQ & 0.6702 & 0.2955 & 1.5547 \\
Gemini 2.5 Pro & Bioactive & Hypothesis & 0.8324 & 0.1947 & \\
Gemini 2.5 Pro & Bioactive & Experimental & 0.7744 & 0.2651 & \\

Gemini 2.5 Pro & Mechanical & Topic+RQ & 0.3398 & 0.2923 & 1.02765 \\
Gemini 2.5 Pro & Mechanical & Hypothesis & 0.5431 & 0.3147 & \\
Gemini 2.5 Pro & Mechanical & Experimental & 0.6293 & 0.2435 & \\

Gemini 2.5 Pro & Nanomaterials & Topic+RQ & 0.5669 & 0.3703 & 1.41445 \\
Gemini 2.5 Pro & Nanomaterials & Hypothesis & 0.7762 & 0.1648 & \\
Gemini 2.5 Pro & Nanomaterials & Experimental & 0.7096 & 0.3049 & \\

Gemini 3.1 Pro & All & Topic+RQ & 0.6033 & 0.3331 & 1.4381 \\
Gemini 3.1 Pro & All & Hypothesis & 0.7545 & 0.2672 & \\
Gemini 3.1 Pro & All & Experimental & 0.7639 & 0.2838 & \\

Gemini 3.1 Pro & Bioactive & Topic+RQ & 0.7837 & 0.2337 & 1.61815 \\
Gemini 3.1 Pro & Bioactive & Hypothesis & 0.8214 & 0.2219 & \\
Gemini 3.1 Pro & Bioactive & Experimental & 0.8098 & 0.3179 & \\

Gemini 3.1 Pro & Mechanical & Topic+RQ & 0.4235 & 0.3100 & 1.29845 \\
Gemini 3.1 Pro & Mechanical & Hypothesis & 0.7241 & 0.2818 & \\
Gemini 3.1 Pro & Mechanical & Experimental & 0.7252 & 0.2448 & \\

Gemini 3.1 Pro & Nanomaterials & Topic+RQ & 0.6028 & 0.3425 & 1.3978 \\
Gemini 3.1 Pro & Nanomaterials & Hypothesis & 0.7181 & 0.2808 & \\
Gemini 3.1 Pro & Nanomaterials & Experimental & 0.7566 & 0.2776 & \\

\bottomrule
\end{longtable}

\end{document}